\documentclass{article}
\usepackage{spconf,graphicx}
\usepackage{amsfonts,amsmath,amssymb,amsthm}
\usepackage{pgfplots}
\usepackage{xcolor}
\usepackage{color,soul}
\usepackage{xspace}
\usepackage[breaklinks=true]{hyperref}

\usepackage[numbers,sort]{natbib}
\usepackage{cleveref}

\newcommand{\ie}{\textit{i}.\textit{e}.}
\newcommand{\eg}{\textit{e}.\textit{g}.}
\def\newpara{\vspace{2pt}}

\title{VGG-Sound: A Large-scale Audio-Visual Dataset}
\name{Honglie Chen, Weidi Xie, Andrea Vedaldi and Andrew Zisserman}
\address{VGG, Department of Engineering Science, University of Oxford, UK \\
\texttt{\{hchen,weidi,vedaldi,az\}@robots.ox.ac.uk}
}

\newcommand{\mypar}[1]{\par\noindent\textbf{#1}~}

\begin{document}
%
\maketitle
\begin{abstract}
Our goal is to collect a large-scale audio-visual dataset with low label
noise from videos `in the wild' using computer vision techniques.
The resulting dataset can be used for training and evaluating audio recognition models.
We make three contributions. 
First, we propose a scalable pipeline based on computer vision techniques to create an audio dataset
from open-source media. 
Our pipeline involves obtaining videos from YouTube; 
using image classification algorithms to localize audio-visual correspondence; 
and filtering out ambient noise using audio verification. 
Second, we use this pipeline to curate the {\tt VGG-Sound} dataset consisting of 
$200$k videos for $309$ audio classes~(detail statistics in Table~\ref{tab:splits}).  
Third, we investigate different architectures along with Convolutional Neural Network~(CNN).
establishing audio recognition baselines for our new dataset.
Compared to existing audio datasets,  
{\tt VGG-Sound} ensures audio-visual correspondence and is collected under unconstrained conditions.
Code and the dataset are available at \url{http://www.robots.ox.ac.uk/~vgg/data/VGGSound/}.
\end{abstract}

\section{Introduction}
\label{sec:intro}

Large-scale datasets~\cite{Everingham10,Deng09} have played a crucial role,
in many deep learning recognition tasks~\cite{Simonyan15,He16,Hershey17}.
In the audio analysis domain, 
while several datasets have been released in the past few years~\cite{Salamon14,Foggia15,Fonseca17,Mesaros16},
the data collection process usually requires extensive human efforts, 
making it unscalable and often limited to narrow domains.
\emph{AudioSet}~\cite{Gemmeke17}, is
a large-scale audio-visual dataset containing over 2 million clips in unconstrained conditions.
This is a valuable dataset, but it required expensive human verification to construct it.
In contrast to these manually curated datasets, recent papers have
demonstrated the possibility of collecting high-quality human speech datasets
in an automated and scalable manner by using computer vision algorithms~\cite{Nagrani17,Nagrani20,Chung18a}. 

In this paper, our objective is to collect a large-scale audio
dataset, similar to \emph{AudioSet}, containing various sounds in the
natural world and obtained `in the wild'  from unconstrained open-source media.
We do this using a pipeline based on computer vision techniques  that guarantees 
audio-visual correspondence (\ie~the sound source is visually evident) and low label noise,
yet requires only minimal manual effort.

\begin{table}[h]
\centering
\small
\begin{tabular}{|c|c|c||c|c|c|}
\hline
Train 	   & Val & Test & Total train & Total & Classes \\ \hline
130-900 & 20         & 50   & 177,837 & 199,467& 309 \\ \hline
\end{tabular}
\caption{{\tt VGG-Sound} Dataset Statistics. The number of clips for each class in the train/val/test partitions and the total numbers of train and classes. 
}
\label{tab:splits}
\end{table}

Our contributions are three-fold:
The first is to propose an automated and scalable pipeline for creating an `in the wild' audio-visual dataset with low label noise.
By using existing image classification algorithms,
our method can generate accurate annotations, circumventing the need for human annotation.
Second, we use this method to curate 
{\tt VGG-Sound}, a large-scale dataset with over $200$k video clips~(visual frames and audio sound) for $309$ audio classes, from  YouTube videos. Each 10s clip contains frames that \emph{show} the object making the sound, and the audio
track contains the the  \emph{sound} of the object. There are at least $200$ clips for each audio class.
Our third contribution is to establish baseline results for audio recognition on this new dataset.
To this end, we investigate different architectures, 
global average pooling and NetVLAD~\cite{Arandjelovic16, Xie19a}, 
for training deep CNNs on spectrograms extracted directly from the audio files with little pre-processing.

We expect {\tt VGG-Sound} to be useful for both audio recognition and audio-visual prediction tasks.
The goal of audio recognition is to determine the semantic content of an acoustic signal, 
\eg~recognizing the sound of a car engine, or a dog barking, \emph{etc}.
In addition,  {\tt VGG-Sound} is equally well suited for studying multi-modal audio-visual analysis tasks,
for example, \emph{audio grounding} aims to localize a sound spatially, 
by identifying in an image the object(s) emitting it~\cite{Arandjelovic17,Kidron05}.
Another important task is to separate the sound of specific objects as they appear in a given frame or video clip~\cite{Owens16, Zhao18}.

\vspace{-2mm}
\section{Related Work}\label{sec:related}
\vspace{-2mm}
\mypar{Audio and audio-visual datasets.}
Several audio datasets exist, as shown in Table~\ref{tab:compare_stat}.
The UrbanSound dataset~\cite{Salamon14} contains more than $8$k urban sound recordings for $10$ classes drawn from the urban sound taxonomy.
The Mivia Audio Events Dataset~\cite{Foggia15} focuses on surveillance applications and contains $6$k audio clips for $3$ classes.
The Detection and Classification of Acoustic Scenes and Events (DCASE) community organizes audio challenges annually,
for example, the authors of~\cite{Mesaros19} released a dataset containing $17$ classes with more than $56$k audio clips.
These datasets are relatively clean, but the scale is often too small to train the data-hungry Deep Neural Networks~(DNNs).

To remedy this shortcoming, a large-scale dataset of video clips was released by Google.
This dataset, 
called \emph{AudioSet}, contains more than $2$ million clips drawn from YouTube and is helpful not only for audio research, 
but audio-visual research as well, where the audio and visual modalities are analysed jointly.
This dataset is a significant milestone,
however, the process used to curate \emph{AudioSet} requires extensive human rating and filtering.
In addition, 
the authors of~\cite{Tian18} manually curated a high-quality, 
but small dataset that guarantees audio-visual correspondence for multi-modal learning, 
where the objects or events that are the cause of a sound must also be observable in the visual stream.


\begin{table}[!htb]
\vspace{-4pt}
\footnotesize
\centering
\begin{tabular}{|c|c|c|c|c|c|c|c|}
\hline
Datasets 			 		& \# Clips & Length & \# Class  & Video 	   & AV-C    \\ \hline
UrbanSound~\cite{Salamon14} & $8$k	   & $8.75$h 	& $10$   	    & $\times$     &  $\times$     \\ 
MIVIA~\cite{Foggia15}		& $6$k     & $29$h   	& $3$    	    & $\times$     &  $\times$      \\ 
DCASE2017~\cite{Mesaros19}	& $57$k    & $89$h      & $17$          & $\times$ 	   & $\times$         \\ 
FSD~\cite{Fonseca17}	    & $24$k    & $119$h     & $398$         & $\times$ 	   & $\times$        \\ 
AudioSet~\cite{Gemmeke17}	& $2.1$m   & $5833$h    & $527$         & $\checkmark$ & $\times$       \\ 
AVE~\cite{Tian18}		    & $4$k     & $11.5$h    & $28$          & $\checkmark$ & $\checkmark$     \\ 
\textbf{VGG-Sound~(Ours)}	& $200$k   & $550$h 	& $309$ 	    & $\checkmark$ & $\checkmark$   \\  \hline
\end{tabular}
\caption{Statistics for recent audio datasets.
``\# Clips'', the number of clips in the dataset; 
``Length'', the total duration of the dataset;
``\# Classes'', number of classes in the dataset;
``Video'', whether videos are available;
``AV-C'', whether audios and videos correspond, 
in the sense that  the sound source is always visually evident within the video clip.}
\label{tab:compare_stat}
\vspace{-2mm}
\end{table}


\mypar{Audio Recognition.}
Audio Recognition, namely the problem of classifying sounds, has traditionally been addressed by means of models such as Gaussian Mixture Models (GMM)~\cite{Zhuang10} and Support Vector Machines (SVM)~\cite{Temko06} trained by using hand-crafted low-dimension features such as the Mel Frequency Cepstrum Coefficients (MFCCs) or i-vectors~\cite{Zhen13}.
However, the performance of MFCCs in audio recognition degrades rapidly in ``unconstrained'' environments that include real-world noise~\cite{Yapanel02,Hansen01}.
More recently, the success of deep learning has motivated approaches based on CNNs~\cite{Naoya16,Hershey17} or RNNs~\cite{Parascandolo16,Choi17,Xu18}.
In this paper, rather than developing complex DNN architectures specific to audio recognition, we choose to illustrate the benefits of our new benchmark dataset by training baselines to serve as comparison for future research.
To this end, we train powerful ResNet architectures 
for audio recognition tasks~\cite{Arandjelovic16, Xie19a}.
\begin{figure*}[t]
\centering
\includegraphics[width=0.95\textwidth]{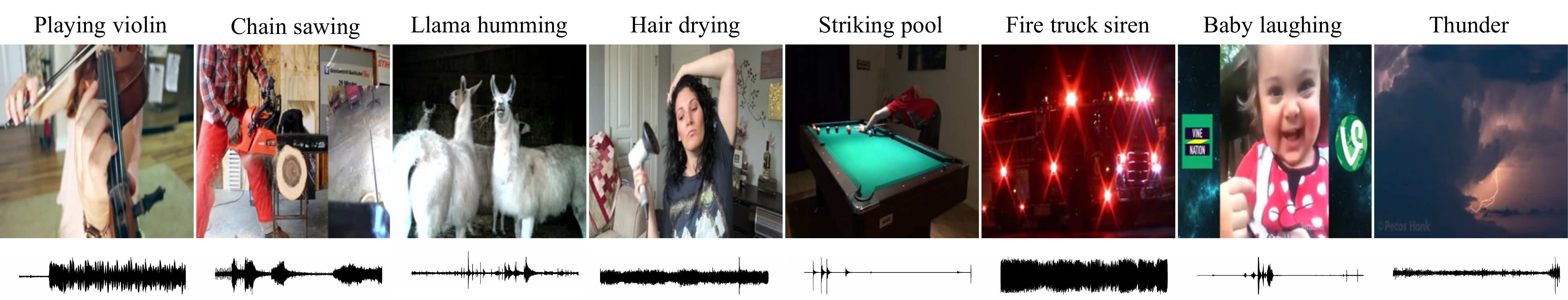}
\vspace{-10pt}
\includegraphics[width=1\textwidth]{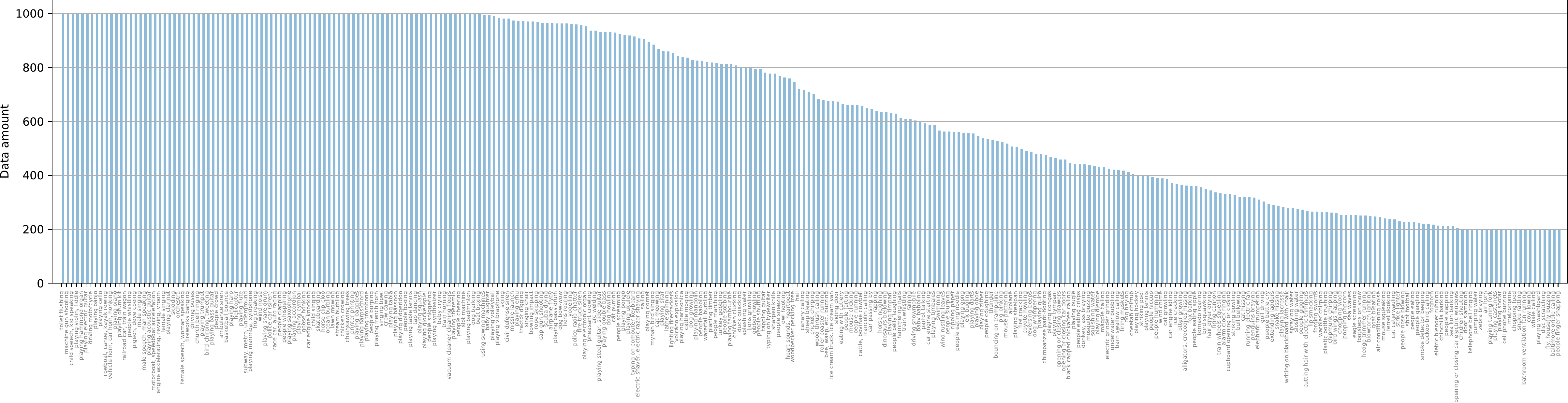}
\caption{The top row in this figure shows example video frame and audio pairs of {\tt VGG-Sound} classes, the bottom bar chart demostrates {\tt VGG-Sound} classes with sizes of each audio class sorted by descending order.}
\vspace{-5mm}
\label{fig:bar}
\end{figure*}

\section{The VGG-Sound dataset}\label{sec:method}
{\tt VGG-Sound} contains over $200$k clips for $309$ different sound classes.
The dataset is audio-visual in the sense that the object that emits each sound is also visible in the corresponding video clip. Figure~\ref{fig:bar} shows example cropped image frames, corresponding audio waveforms,  
and a histogram detailing the statistics for each class.
Each sound class contains $200$--$1000$ $10$s clips, with no more than $2$ clips per video.
The set of sound labels is flat (\ie~there is no hierarchy as in \emph{AudioSet}).
Sound classes can be loosely grouped as: people, animals, music, sports, nature, vehicle, home, tools, and others.
All clips in the dataset are extracted from videos downloaded from YouTube,
spanning a large number of challenging acoustic environments and noise characteristics of real applications.

In the following sections, 
we describe the multi-stage approach that we have developed to collect the dataset. 
The process can be described as a cascade that starts from a large number of potential audio-visual classes and corresponding videos, and then progressively filters out classes and videos to leave a smaller number of clips that are annotated reliably.
The number of classes and videos after each stage of this process is shown in Table~\ref{tab:stat_after_steps}.
The process is extremely scalable and
only requires manual input at a few points for well defined tasks.

\begin{table}[!htb]
\vspace{1mm}
\begin{center}
\small
 \begin{tabular}{|c| c |c |c|} 
 \hline
 Stages & Goal & \# Classes & \# Videos \\ [0.5ex] 
 \hline\hline
 1 & Candidate videos & $600$ &  $1$m \\ 
 \hline
 2 & Visual verification & $470$ & $550$k \\
 \hline
 3 & Audio verification & $390$ & $260$k \\
 \hline
 4 &  Iterative noise filtering & $309$ & $200$k \\
 \hline
\end{tabular}
\caption{The number of classes and videos after each stage of the generation pipeline.
Note, classes with less than $100$ videos are removed from the dataset.}
\label{tab:stat_after_steps}
\vspace{-5mm}
\end{center}
\end{table}

\newpara\noindent\textbf{Stage 1: Obtaining the class list and candidate videos.}
The first step is to determine a tentative list of sound classes to include in the dataset.
We follow \emph{three} guiding principles in order to generate this list.
First, the sounds should be \emph{in the wild}, in the sense that they should occur in real life scenarios,
as opposed to artificial sound effects.
Second, it must be possible to \emph{ground and verify the sounds visually}.
In other words, our sound classes should have a clear visual connotation too, 
in the sense of being predictable with reasonable accuracy from images alone.
For instance, the sound `electric guitar' is visually recognizable as it is generally possible to visually recognize someone playing a guitar, but `happy song' and `pop music' are not: these classes are too abstract for visual recognition and so they are not included in the dataset.
Third, the classes should be \emph{mutually exclusive}.
Although we initialize the list using the label hierarchies in existing audio datasets~\cite{Gemmeke17,Foggia15,Fonseca17} 
and other hierarchical on-line sources, our classes are only leaf nodes in these hierarchies.
In this manner, the label set in {\tt VGG-Sound} is flat and contains only one label for each clip.
For instance, if the clip contains the sound of a car engine, 
then the label will  be only ``car engine''; the more general class ``engine'' is not included in the list.

The initial list of classes, constructed in this manner, had $600$ items.
Each class name is used as a search query for YouTube to automatically download corresponding candidate videos.
In order to increase the chance of obtaining relevant videos, the class names are further transformed to generate variants of each textual query as follows:
(1) forming `verb+(ing) object' sentences, \eg~`playing electric guitar',`ringing church bells', \emph{etc.}
(2) submitting the query after translation to different languages, as is done in~\cite{Carreira18},  such as English, Spanish and Chinese, etc;
(3) adding possible synonym phrase which specify the same sounds, \eg~`steam hissing: water boiling, liquid boiling, \emph{etc}.'
In total, over $1$m videos were downloaded from YouTube in this manner.

\newpara\noindent\textbf{Stage 2: Visual verification.}
The purpose of this stage is to verify and localize the visual signature in the downloaded videos.
In detail, for each {\tt VGG-Sound} class,
the corresponding visual signature is given by image classifiers.
For example, `playing violin' and `cat meowing'~in {\tt VGG-Sound} can be matched directly to 
the OpenImage classifiers~\cite{Openimages} `violin' and `cat'. 
These associations are proposed automatically by matching keywords and then verified manually.

However, half of {\tt VGG-Sound} classes (\eg~`hail', 'playing ukulele') could not be matched directly 
to OpenImage classifiers  in this manner.
To tackle this issue, we relax the way sound labels are matched to visual labels via semantic word embeddings.
Specifically, we 
convert our $600$ sound classes and the $5000$ OpenImage classes to word2vec embeddings~\cite{Mikolov13b}.
These embedding have $512$ dimensions, so this step results in matrices 
$S \in \mathcal{R}^{600 \times 512}$ and $O \in \mathcal{R}^{5000 \times 512}$, respectively for sound and image labels.
We then compute the cosine similarity between the two matrices, 
resulting in an affinity matrix $A \in \mathcal{R}^{600 \times 5000} = S O^\top$ that represents the strength of the similarity between sound and image classes.
The top $20$ OpenImage classes for each of the $600$ sound classes are then selected as the visual signature of the corresponding sound.
For example, `hail' was matched to `nature, nature reserve, rain and snow mixed, lightning, thunderstorm, \emph{etc}.' and
`playing electric guitar' to `electric guitar, guitar, acoustic-electric guitar, musical instrument, \emph{etc}'.
%
%
%
%
\begin{table*}[!htb]
\centering
\footnotesize
\begin{tabular}{|c|c|c|c|c|c|c|c|}
\hline
  & Model 	 & Test      & mAP     & AUC     & d-prime & Top1 acc   & Top5 acc  \\ \hline
A & ResNet18 & ASTest    & $0.404$ & $0.944$ & $2.253$ & 0.404      & 0.679\\ \hline
B & ResNet34 & ASTest    & $0.409$ & $0.947$ & $2.292$ & 0.410      & 0.691\\ \hline

C & ResNet50 & ASTest    & $0.412$ & $0.949$ & $2.309$ & 0.415      & 0.698 \\ \hline
D & ResNet18 & VGG-Sound & $0.516$ & $0.968$ & $2.627$ & 0.488      & 0.746 \\ \hline

E & ResNet34 & VGG-Sound & $0.529$ & $0.972$ & $2.703$ & 0.505      & 0.758 \\ \hline

F & ResNet50 & VGG-Sound & $0.532$ & $0.973$ & $2.735$ & 0.510      & 0.764 \\ \hline
\end{tabular}
\caption{We compare the results using various combination of architectures and test sets.
All models here are trained using VGG-Sound training set.
``ASTest'' is the intersection of the AudioSet and VGG-Sound testsets,
``Top1 acc'' and ``Top5 acc'' refers  to the top 1 and top 5 accuracy, respectively.
}
\vspace{-5mm}
\label{tab:results}
\end{table*}

After determining these associations, the OpenImage pre-trained classifier are run on the downloaded videos,
and  the  $10$ frames in the video that receive the highest prediction score are selected, 
provided the score is above an absolute confidence threshold of $0.2$.
The frames that pass this test are assumed to contain the visual content selected by the classifier.
Clips are then created by taking $5$ seconds at either side of these representative frames.
After this stage, the number of sound classes is reduced from the original $600$ to $470$, 
due either an initial scarcity of potential video matches or by failed visual verification.

\newpara\noindent\textbf{Stage 3. Audio verification to remove negative clips.}
Despite visual verification, our clips are still not guaranteed to contain the desired sound, 
as an object being visible does not imply that it emits a sound at the same time;
in fact, we found that many clips where the correct object was in focus, 
contained instead generic sounds from humans, 
such as a narrator describing an image or video, or background music.
Since these issues are fairly specific, 
we address them by finetuning the VGGish model with only three sound classes: speech, music and others.
The finetuned classifier is typically reliable as most of the existing datasets offers higly clean data of these classes.
We use it to reject clips.
For example, using a threshold $0.5$, in `playing bass guitar' videos, 
we reject any clip for which ``speech'' is greater than the threshold, but allow music; 
while for `dog barking' videos, both speech and music are rejected.
After this stage, there are $390$ classes left with at least $200$ validated video clips.
Note that our selection process aims to reject ``false positive'', 
\ie~inappropriate sounds in each class,
we do not attempt to use an audio classifier to select positive clips as that risks losing hard positive audio samples.

\newpara\noindent\textbf{Stage 4: Iterative noise filtering.}
For this final clean up stage of the process,  $20$ video clips are randomly sampled from each class
and manually checked (both visually and on audio) that they belong to the class.
Classes with at least $50\%$ correct are kept and the other classes are discarded.
The total set of video clips remaining forms our candidate dataset.
Note that, at this stage, 
the candidate videos can be categorized by audio as one of three types: (i) 
audios that are clearly of the correct category, \emph{i.e.}~easy positives; (ii)
audios of the correct category, but with a mixture of sounds, \emph{i.e.}~hard positives; 
or (iii) incorrect audios, \emph{i.e.}~false positive.

To further curate the candidate dataset, 
we make three assumptions: 
First, there is no systematic bias in the noisy samples, 
by that we mean, the false positives are not from the same category.
Second, Deep CNNs tend to end up with different local minimas and prediction errors, 
ensembling different models can therefore result in a prediction that is both more stable 
and often better than the predictions of any individual member model.
Third, when objects emit sound, there exists particular visual patterns, 
\emph{e.g.}~a ``chimpanzee pant-hooting'' will mostly happen with moving bodies.

Exploiting the first two assumptions, the videos of each class are randomly divided into two sets,
and an audio classifier is trained on half the candidate videos and used to predict the class of the other half. 
This process is done twice so that each clip has $2$ predictions. 
To obtain relatively easy and precise postives, 
we keep the clips whose actual class-label falls into the top-$3$ of the predictions from the ensembled models.
In order to mine the harder positives, 
we exploit the third assumption by computing visual features for the positive clips,
and perform visual retrieval from the rest of data that has been rejected by the audio classifiers.
Using a visual classifier can result in similar looking  visual clips but disparate hard-positive audio clips.
Lastly, we train a new audio classifier~(ResNet18) with all easy and hard clips, 
and retrieve more data from that  rejected so far.
This increases the number of video clips and forms our final dataset:
{\tt VGG-Sound} with $309$ classes of over $200$k videos, 
and each class contains $200$--$1000$ audio-visual corresponding clips.
Note, we did a deduplication process to remove repeated uploaded clips with different YouTube IDs. This is done by removing the ones with same visual representations.

\section{Experiments}\label{sec:experiment}

\subsection{Experimental Setup and Evaluation}

\mypar{Experimental setup.}
We investigate the audio recognition task on 
our new {\tt VGG-Sound} dataset.
At training time, we train models using VGG-Sound training set.
At testing time, we take the intersection of \emph{AudioSet} and {\tt VGG-Sound} to form a single testset called AStest (and remove any videos in AStest that are in the training sets of  \emph{AudioSet} or {\tt VGG-Sound}).
This leads $164$ classes and $7$k clips in AStest. 
We then test on both the AStest and VGG-Sound test set.



\vspace{5pt}
\mypar{Evaluation Metrics.}
We adopt the evaluation metrics of~\cite{Hershey17}, \ie~mean average precision (mAP),
AUC,  and equivalent d-prime class separation.

\subsection{Implementation Details}
During training,
we randomly sample $5$s from the $10$s audio clip and apply a short-time Fourier transform on the  sample,
resulting a $257\times500$ spectrogram.
During testing, we directly feed the $10$s audio into the network.

All experiments were trained using the Adam optimizer with cross entropy loss.
The learning rate starts with $10^{-3}$ and is reduced by a factor of 10 after training plateaus.
Since we define classes with only leaf node, we use softmax layer in the last layer.


\subsection{Results}
We set up baselines for audio classification on the {\tt VGG-Sound} dataset.
From the experimental results in Table~\ref{tab:results}, 
we can draw the following conclusions:
A comparison of~Model-A, Model-B and~Model-C, demonstrates that 
a  model trained with deeper architectures clearly works better than one  trained with a shallow architecture.
Testing on the full {\tt VGG-Sound} test set~Model-D shows a better
result comparing to the one testing only on ASTest~Model-A. This is
because ASTest only contains AudioSet testing clips and these  tend to have
multiple sounds. These are hard examples for a model to predict, which
can also be seen from the Top5 accuracy as the gap between~Model-D
and~Model-A is largely reduced.
\vspace{-2mm}
\section{Conclusion}
\label{sec:conclusion}
\vspace{-2mm}
In this paper, we propose an automated pipeline for collecting a large-scale audio-visual dataset -- {\tt VGG-Sound}, 
which contains more than $200$k videos and $309$ classes for  ``unconstrained'' conditions.
We also compare CNN architectures 
to provide baseline results for audio recognition on {\tt VGG-Sound}.

\newpara\noindent\textbf{Acknowledgement.}
Financial support was provided by the EPSRC Programme Grant Seebibyte EP/M013774/1.

{\small
\bibliographystyle{IEEEbib}
\bibliography{shortstrings,vgg_local,vgg_other}}

\appendix
\vspace{-2mm}
\section{List of VGG-Sound Classes}
\vspace{-2mm}
\label{appendix:a}

\begin{enumerate}
\item ambulance siren (1050) (common)
\item basketball bounce (1050) (common)
\item bird chirping, tweeting (1050) (common)
\item chainsawing trees (1050) (common)
\item chicken crowing (1050) (common)
\item child speech, kid speaking (1050) (common)
\item driving buses (1050) (common)
\item driving motorcycle (1050) (common)
\item engine accelerating, revving, vroom (1050) (common)
\item female singing (1050) (common)
\item female speech, woman speaking (1050) (common)
\item fireworks banging (1050) (common)
\item helicopter (1050) (common)
\item male singing (1050) (common)
\item male speech, man speaking (1050) (common)
\item motorboat, speedboat acceleration (1050) (common)
\item orchestra (1050) (common)
\item people booing (1050) (common)
\item people crowd (1050) (common)
\item pigeon, dove cooing (1050) (common)
\item playing accordion (1050) (common)
\item playing acoustic guitar (1050) (common)
\item playing banjo (1050) (common)
\item playing bass guitar (1050) (common)
\item playing bassoon (1050) (common)
\item playing cello (1050) (common)
\item playing cymbal (1050) (common)
\item playing didgeridoo (1050) (common)
\item playing drum kit (1050) (common)
\item playing electric guitar (1050) (common)
\item playing flute (1050) (common)
\item playing harp (1050) (common)
\item playing marimba, xylophone (1050) (common)
\item playing piano (1050) (common)
\item playing saxophone (1050) (common)
\item playing trombone (1050) (common)
\item playing violin, fiddle (1050) (common)
\item police car (siren) (1050) (common)
\item race car, auto racing (1050) (common)
\item railroad car, train wagon (1050) (common)
\item rowboat, canoe, kayak rowing (1050) (common)
\item singing bowl (1050) (common)
\item tap dancing (1050)
\item toilet flushing (1050) (common)
\item vehicle horn, car horn, honking (1050) (common)
\item people burping (1049) (common)
\item playing clarinet (1049) (common)
\item playing hammond organ (1049) (common)
\item playing squash (1049)
\item playing table tennis (1049)
\item playing tabla (1048) (common)
\item playing glockenspiel (1046) (common)
\item playing harpsichord (1022) (common)
\item frog croaking (1018) (common)
\item vacuum cleaner cleaning floors (1009) (common)
\item playing theremin (1007) (common)
\item playing badminton (1000)
\item slot machine (1000)
\item subway, metro, underground (1000) (common)
\item beat boxing (998)
\item child singing (994) (common)
\item crow cawing (993) (common)
\item playing volleyball (991)
\item car engine knocking (990) (common)
\item playing snare drum (987) (common)
\item machine gun shooting (986) (common)
\item ocean burbling (983) (common)
\item skiing (980)
\item civil defense siren (979)
\item missile launch (974)
\item playing erhu (971)
\item tractor digging (970)
\item bowling impact (969)
\item rope skipping (967)
\item airplane flyby (964)
\item lions roaring (964) (common)
\item scuba diving (963)
\item yodelling (961)
\item playing bagpipes (960) (common)
\item fire truck siren (958) (common)
\item police radio chatter (958)
\item pheasant crowing (937)
\item arc welding (934)
\item dog howling (929) (common)
\item playing steel guitar, slide guitar (929) (common)
\item wind noise (922) (common)
\item playing bongo (920)
\item skateboarding (914) (common)
\item playing synthesizer (909) (common)
\item electric shaver, electric razor shaving (908)
\item playing ukulele (907) (common)
\item skidding (906) (common)
\item cap gun shooting (904) (common)
\item people clapping (903) (common)
\item playing cornet (893) (common)
\item people sniggering (891) (common)
\item playing vibraphone (885) (common)
\item mynah bird singing (884)
\item people whistling (883) (common)
\item horse clip-clop (882) (common)
\item baby laughter (881) (common)
\item printer printing (881) (common)
\item playing french horn (865) (common)
\item playing bass drum (863) (common)
\item cat purring (861) (common)
\item playing sitar (860) (common)
\item lathe spinning (859)
\item lawn mowing (858) (common)
\item lighting firecrackers (854) (common)
\item volcano explosion (842)
\item playing electronic organ (839) (common)
\item playing trumpet (835) (common)
\item dog growling (826) (common)
\item playing harmonica (824) (common)
\item playing mandolin (824) (common)
\item people marching (817)
\item planing timber (816)
\item playing tambourine (811)
\item playing double bass (810) (common)
\item church bell ringing (807) (common)
\item gibbon howling (792)
\item goose honking (788) (common)
\item tapping guitar (780) (common)
\item sharpen knife (776)
\item train horning (771) (common)
\item dog barking (767) (common)
\item singing choir (762) (common)
\item heart sounds, heartbeat (756)
\item owl hooting (753) (common)
\item people whispering (747) (common)
\item typing on typewriter (745) (common)
\item woodpecker pecking tree (745)
\item cattle mooing (735) (common)
\item people screaming (731) (common)
\item hail (718)
\item canary calling (717)
\item people cheering (704) (common)
\item cricket chirping (702) (common)
\item stream burbling (690) (common)
\item people sneezing (688) (common)
\item wood thrush calling (681)
\item roller coaster running (678)
\item ice cream truck, ice cream van (676)
\item people shuffling (675) (common)
\item baby crying (667) (common)
\item mouse clicking (660)
\item francolin calling (651)
\item turkey gobbling (646) (common)
\item cattle, bovinae cowbell (644) (common)
\item rapping (638) (common)
\item dinosaurs bellowing (633)
\item playing timpani (629) (common)
\item dog bow-wow (628) (common)
\item using sewing machines (626) (common)
\item chicken clucking (625) (common)
\item splashing water (618) (common)
\item hammering nails (613)
\item train whistling (608)
\item driving snowmobile (603)
\item duck quacking (602) (common)
\item lions growling (599) (common)
\item baby babbling (598) (common)
\item car passing by (595) (common)
\item parrot talking (593)
\item sailing (592) (common)
\item cat meowing (588) (common)
\item horse neighing (587) (common)
\item people babbling (587) (common)
\item playing timbales (587)
\item sliding door (587) (common)
\item swimming (566)
\item wind rustling leaves (563) (common)
\item people slurping (562)
\item typing on computer keyboard (561) (common)
\item people eating noodle (560)
\item ripping paper (560)
\item elk bugling (557)
\item playing gong (557) (common)
\item playing darts (555)
\item people farting (554) (common)
\item bee, wasp, etc. buzzing (551) (common)
\item playing oboe (546) (common)
\item playing zither (538) (common)
\item eating with cutlery (528) (common)
\item bouncing on trampoline (526)
\item people belly laughing (517) (common)
\item sheep bleating (515) (common)
\item waterfall burbling (514) (common)
\item airplane (506)
\item playing steelpan (504) (common)
\item fire crackling (492) (common)
\item car engine starting (491) (common)
\item coyote howling (490)
\item reversing beeps (487) (common)
\item people sobbing (481) (common)
\item playing guiro (477)
\item chimpanzee pant-hooting (473)
\item raining (473) (common)
\item playing congas (465)
\item playing tympani (463)
\item dog whimpering (461) (common)
\item opening or closing car doors (459)
\item opening or closing drawers (458) (common)
\item people coughing (453) (common)
\item black capped chickadee calling (446)
\item pig oinking (444) (common)
\item people eating crisps (442)
\item playing bugle (442)
\item donkey, ass braying (438) (common)
\item mosquito buzzing (438) (common)
\item squishing water (435)
\item mouse pattering (430) (common)
\item playing djembe (430)
\item magpie calling (429)
\item electric grinder grinding (425)
\item people running (422) (common)
\item barn swallow calling (420)
\item underwater bubbling (420)
\item firing muskets (417)
\item dog baying (412) (common)
\item cheetah chirrup (403)
\item playing hockey (402)
\item striking pool (401)
\item people hiccup (394)
\item people humming (391)
\item playing tennis (390)
\item wind chime (389)
\item cat growling (387)
\item car engine idling (370)
\item ice cracking (365)
\item otter growling (363)
\item alligators, crocodiles hissing (361)
\item fox barking (361)
\item people eating apple (360)
\item tornado roaring (356)
\item bird squawking (348) (common)
\item hair dryer drying (343) (common)
\item firing cannon (336) (common)
\item train wheels squealing (333) (common)
\item alarm clock ringing (331)
\item cupboard opening or closing (330)
\item striking bowling (326)
\item thunder (325) (common)
\item bull bellowing (320)
\item cat hissing (320)
\item penguins braying (318)
\item running electric fan (318)
\item elephant trumpeting (310)
\item golf driving (303)
\item people battle cry (291)
\item snake hissing (285)
\item playing lacrosse (282)
\item writing on blackboard with chalk (280)
\item spraying water (278)
\item sloshing water (276)
\item people giggling (273)
\item cutting hair with electric trimmers (268)
\item lip smacking (265)
\item forging swords (264)
\item plastic bottle crushing (264)
\item warbler chirping (264)
\item chipmunk chirping (261)
\item bird wings flapping (258) (common)
\item chopping wood (254)
\item popping popcorn (253)
\item eagle screaming (252)
\item footsteps on snow (251)
\item hedge trimmer running (251)
\item blowtorch igniting (250)
\item sea waves (250)
\item goat bleating (246) (common)
\item air conditioning noise (245)
\item mouse squeaking (240) (common)
\item ferret dooking (239)
\item cat caterwauling (237) (common)
\item strike lighter (229)
\item people nose blowing (228)
\item people eating (226) (common)
\item shot football (226)
\item people gargling (221)
\item smoke detector beeping (221)
\item cuckoo bird calling (218)
\item foghorn (217)
\item eletric blender running (214)
\item chinchilla barking (213)
\item people slapping (211)
\item sea lion barking (211)
\item opening or closing car electric windows (205)
\item children shouting (201)
\item baltimore oriole calling (200)
\item bathroom ventilation fan running (200)
\item cell phone buzzing (200)
\item chopping food (200)
\item disc scratching (200)
\item door slamming (200) (common)
\item fly, housefly buzzing (200)
\item metronome (200)
\item people finger snapping (200) (common)
\item playing shofar (200)
\item playing tuning fork (200)
\item playing washboard (200)
\item telephone bell ringing (200)
\item whale calling (200)
\item cow lowing (199)
\item playing castanets (199)
\item pumping water (199) (common)
\item snake rattling (199)
\item zebra braying (199)
\item air horn (197)

\end{enumerate}

\end{document}